\documentclass[sigconf]{acmart}

\AtBeginDocument{%
  \providecommand\BibTeX{{%
    \normalfont B\kern-0.5em{\scshape i\kern-0.25em b}\kern-0.8em\TeX}}}

\setcopyright{acmcopyright}
\copyrightyear{2022}
\acmYear{2022}
\acmDOI{XXXXXXX.XXXXXXX}

\acmConference[ACMMM '22]{ACM Multimedia 2022}{October 10--14,
  2022}{Lisbon, Portugal}
\acmPrice{15.00}
\acmISBN{978-1-4503-XXXX-X/18/06}

\usepackage{soul}
\usepackage{hyperref}
\hypersetup{urlcolor=magenta}

\newcommand{\etal}{\textit{et al}. }

\newcommand{\swrigrf}{\textit{ForcePose}}
\newcommand{\newloss}{gated-MSE}
\newcommand{\newlossshort}{our loss}
\newcommand{\newmetric}{mean $k$-peaks}

\acmSubmissionID{2889}

\begin{document}

\title{Learning to Estimate External Forces of Human Motion in Video}

\author{Nathan Louis}
\authornote{Completed as part of internship at Southwest Research Institute}
\email{natlouis@umich.edu}
\orcid{0000-0003-4502-6012}
\affiliation{%
  \institution{University of Michigan}
  \city{Ann Arbor}
  \state{Michigan}
  \country{USA}
  \postcode{48109}
}

\author{Jason J. Corso}
\affiliation{%
  \institution{University of Michigan}
  \city{Ann Arbor}
  \state{Michigan}
  \country{USA}
  \postcode{48109}}
\email{jjcorso@umich.edu}

\author{Tylan N. Templin}
\affiliation{%
  \institution{Southwest Research Institute}
  \city{San Antonio}
  \state{Texas}
  \country{USA}}
\email{tylan.templin@swri.org}

\author{Travis D. Eliason}
\affiliation{%
  \institution{Southwest Research Institute}
  \city{San Antonio}
  \state{Texas}
  \country{USA}}
\email{travis.eliason@swri.org}

\author{Daniel P. Nicolella}
\affiliation{%
  \institution{Southwest Research Institute}
  \city{San Antonio}
  \state{Texas}
  \country{USA}}
\email{dnicolella@swri.org}

\renewcommand{\shortauthors}{Louis, et al.}

\begin{abstract}
  Analyzing sports performance or preventing injuries requires capturing ground reaction forces (GRFs) exerted by the human body during certain movements. 
Standard practice uses physical markers paired with force plates in a controlled environment, but this is marred by high costs, lengthy implementation time, and variance in repeat experiments; hence, we propose GRF inference from video.  
While recent work has used LSTMs to estimate GRFs from 2D viewpoints, these can be limited in their modeling and representation capacity.
First, we propose using a transformer architecture to tackle the GRF from video task, being the first to do so. 
Then we introduce a new loss to minimize high impact peaks in regressed curves.
We also show that pre-training and multi-task learning on 2D-to-3D human pose estimation improves generalization to unseen motions. And pre-training on this different task provides good initial weights when finetuning on smaller (rarer) GRF datasets.
We evaluate on LAAS Parkour and a newly collected \swrigrf\ dataset; we show up to $19\%$ decrease in error compared to prior approaches.
\end{abstract}


\begin{teaserfigure}
  \includegraphics[width=\textwidth]{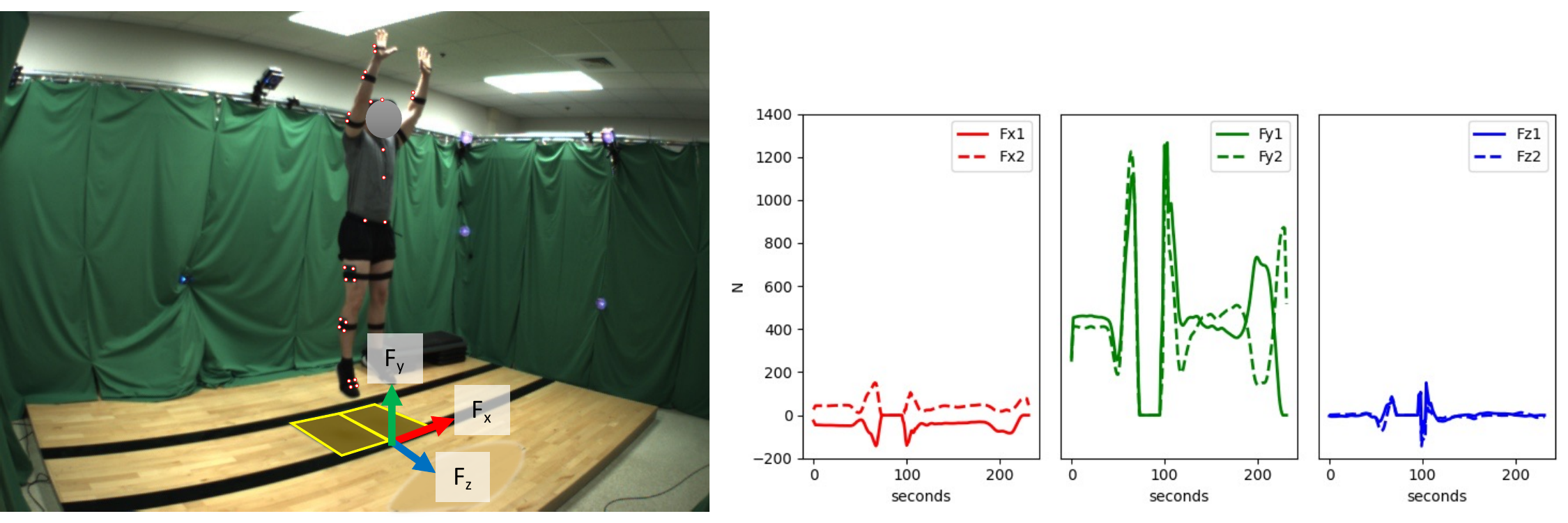}
  \caption{Standard practice for modeling human motion with forces requires physical reflective markers to capture kinematics and force plates to measure GRFs. In contrast, we propose a video-only approach that yields comparable performance yet does not require any physical apparatus to predict forces. Here the subject is performing a jumping movement on two force plates, measured forces are shown on the right. (Reflective markers are highlighted for visibility)}
  \Description{TBD}
  \label{fig:teaser}
\end{teaserfigure}

\maketitle

\section{Introduction}

\label{sec:intro}
For identifying factors in performance improvement or injury prevention of certain physical activities (e.g. running and jumping), an essential quantitative measure is the Ground Reaction Force (GRF) \cite{verheul2019whole}. Used primarily in the field of biomechanics, GRF measures the three-dimensional contact force between the human body and the ground. When paired with calculated kinematics, GRFs can be used to compute inverse dynamics and estimate internal forces and interactions between the different joints, muscles, and bones within a subject \cite{verheul2019whole,zajac1989determining}.

Traditionally \cite{johnson2018predicting,ren2008whole} this process requires a multi-camera marker-based tracking system and force plates in a controlled lab environment as shown in Fig. \ref{fig:teaser}. 
Consequently, this time-consuming and expensive process cannot be replicated outside of a lab environment \cite{karatsidis2017estimation,ren2008whole}.
However, advances in human pose estimation in 2D images and video has opened the doors for marker-less systems to replace this \textit{antiquated} approach, like Morris \etal \cite{morris2021predicting} who use detected keypoints and an LSTM network to predict GRFs directly from video.
But, this work only experiments with a single lateral side-stepping maneuver, includes no implicit or explicit 3D representations, and is limited by the modeling and representation capacity of the LSTM. LSTMs suffer from vanishing gradients on really long sequences and lack the capabilities of modern attention-based models.

In contrast, we use a transformer architecture to predict GRFs from video.
Transformer networks \cite{vaswani2017attention} originated as a driving force in natural language processing (NLP)  because of their computational efficiency and ability to learn longer range dependencies than LSTM and RNN networks.
But they typically require orders of magnitudes more training data to generalize well \cite{dosovitskiy2020image}. 
We address this by leveraging features learned from 2D-to-3D human pose estimation (HPE) via pre-training and multi-task learning as a subtask, learning to infer a 3D pose from a 2D input. 
Because GRFs are predicted on a three-dimensional plane, subtle out-of-plane movements from a 2D view maybe harder to capture with a restrictive point-of-view. So we encode an approximate 3D representation from a 2D pose, along with context of the input sequence, as a joint feature representation transferable to the target force prediction task.
Pre-training follows the train and finetune paradigm, while multi-task learning optimizes both the 3D pose estimation and force prediction simultaneously.

The typical optimization recipe for GRF prediction uses mean-squared error (MSE) as a loss and root-mean squared error (RMSE) as a metric, but this overlooks a vital portion of the end task.
For load analysis from motion, we care about the peak impact forces and how closely the predicted magnitude matches the ground truth.
For example, an athlete is likelier to incur an injury during more active parts of a movements rather than standing still.
While MSE and RMSE provide a good approximation for best curve fit, they unfairly weigh all parts of the force output.
In periods of relatively no motion from the user, such as right before a jump, the predicted and ground truth forces will be easy to predict---relatively flat values.
These ``low activity'' regions, in the learning process, often cause a muted (underestimated) approximation of peak impact forces to satisfy the flatter portions of the force curves.
To account for this, we introduce a new loss function, \newloss, that prioritizes higher impact regions of the force prediction during training.
We find that this leads to a small cost in RMSE but significantly decreases peak impact errors.

Last, we introduce \swrigrf, a novel dataset to address the GRF prediction problem.
There is an absence of publicly available data for this task, with LAAS Parkour \cite{li2019estimating} a single-view dataset of $28$ videos only, being publicly available.
In comparison, \swrigrf\ is a multi-view dataset with eight subjects performing five movements for multiple trials, resulting in over $1,300$ videos with paired force plate data.
We show that for extremely small datasets, such as LAAS Parkour, the 2D-to-3D pre-training provides up to $19\%$ decrease in error compared to training from random weights. For small-to-medium-sized multiview datasets, such as \swrigrf, employing a multi-task learning regime provides a moderate increase in performance when evaluating on the target task and on unseen motions.

Our main contributions are as follows:
\begin{itemize}
    \item We introduce \newloss\ as a new loss to minimize peak impact errors in GRF prediction.
    \item We show, using a transformer architecture, that pre-training and multi-task learning on 2D-to-3D human pose estimation:
        \begin{itemize}
            \item Provides good initial weights for finetuning on small datasets;
            \item Learns a useful representation for generalizing to novel motions.
        \end{itemize}
    \item We provide the \swrigrf\ dataset, a novel collection of multi-view tracked human motions with time-synchronized force plate data.
\end{itemize}

\section{Related Work}

\subsection{2D and 3D Human Pose Estimation}
The majority of work in Human Pose Estimation (HPE) exists in the 2D space, with most
techniques either in the bottom-up \cite{raaj2019efficient,iqbal2017posetrack} and top-down \cite{xiao2018simple,sun2019deep,wang2020combining} categories, with the latter showing superior performance.
With the increase in performance on images in-the-wild, publicly available frameworks such as OpenPose \cite{cao2019openpose} and Detectron \cite{wu2019detectron2} are directly used out-of-the-box as an intermediate step to many pose detection problems. In this work, we take 2D poses for granted and focus on downstream tasks using these fixed pose detections.
The 3D HPE task features higher-dimensional data that typically requires a motion capture system to create datasets from real scenes \cite{ionescu2013human3}.
Solutions in this field can be categorized into direct 3D inference \cite{moon2020i2l,xu2020ghum,pavlakos2018ordinal}, which regresses a 3D pose directly from pixels, or 2D-to-3D inference \cite{wang2020motion,liu2020attention}, which uses intermediate representations such as a 2D pose detection to produce a 3D pose.
In this work, we argue that 2D-to-3D inference serves as appropriate pre-training and simultaneous supervision for our GRF prediction task.
We postulate that the learned implicit 3D representation will serve as a significant benefit to estimating GRFs.

\subsection{Biomechanical Load Analysis}
Analyzing the biomechanics of human motion requires computed kinematics, traditionally from reflective surface markers, and forces, conventionally from time-synchronized force plates, constrained to a lab environment. 
Existing works infer kinematics using physical markers \cite{ren2008whole,johnson2018predicting,johnson2018predictingcnn} or sensors such as IMUs and accelerometers \cite{alcantara2021predicting,karatsidis2017estimation}.
But the variance in the placement of these physical devices impacts repeatability and results \cite{mcginley2009reliability}, heavily relying on the technical skills of the person placing the markers. Even the attachment of sensors are susceptible to variance, noise and movement artifacts. Hence we use a marker-less (human pose estimation) system to predict marker positions and shift to a neural network approach for predicting exerted forces.

Like our work, Morris \etal \cite{morris2021predicting} move towards a human pose detection using an LSTM on sequences of 2D pose detections to predict GRFs in sidestepping maneuvers.
The authors note a higher error in the medio-lateral (side-to-side) movements which we hypothesize is a result of not integrating poses from multiple views or implicitly utilizing a 3D pose.
In contrast to prior approaches, we are the first to use a transformer to address this problem, leveraging features learned from 2D-to-3D HPE domain to address limitations in single viewpoints.
Given the lack of publicly available data with ground truth contact forces, we introduce a manually collected dataset, \swrigrf\, in our experiments. This dataset contains more dynamic movements and number of data samples in comparison to unpublished datasets \cite{morris2021predicting,jeong2020estimation,strutzenberger2021assessment} described in other works.

\subsection{Temporal Sequence Models}

LSTM and RNN networks have been the defacto choice for temporal sequences and NLP tasks, but transfomer networks \cite{vaswani2017attention} have since dominated the NLP space and more recently vision and language modalities \cite{sun2019videobert,lu2019vilbert,dosovitskiy2020image,carion2020end,bertasius2021space}. A transformer is an attention-based deep learning model that uses multi-headed attention modules to learn relations between parts of an input sequence, without the use of recurrence. 
Compared to LSTM networks, transformers avoid issues such as backpropagating through time and vanishing gradients pitfalls.
But despite the notable developments, transformer networks lack inductive biases and must often train on larger datasets to generalize well \cite{dosovitskiy2020image}. Examples are seen with BERT \cite{devlin2018bert}, training on English Wikipedia (2,500M words), and ViT \cite{dosovitskiy2020image}, training on JFT (300M images).
Recently, transformers have also been introduced in the 2D \cite{snower202015,li2021pose,yang2021transpose} and 3D \cite{ma2021transfusion} HPE tasks. 
Zheng \etal \cite{zheng20213d} presents PoseFormer, which argues that learning separate spatial (intra-joint) and temporal (inter-pose) encodings in two transformers are important for 2D-to-3D pose estimation. We use this transformer architecture as the base transformer in our experiments.
Not only does this include an attention mechanism, lacking in the LSTM network, but it also serves a dual role in our 2D-to-3D HPE subtask.

\section{Data} \label{sec:data}
\subsection{ForcePose Dataset}
\begin{figure*}[t]
    \centering
    \includegraphics[width=\linewidth]{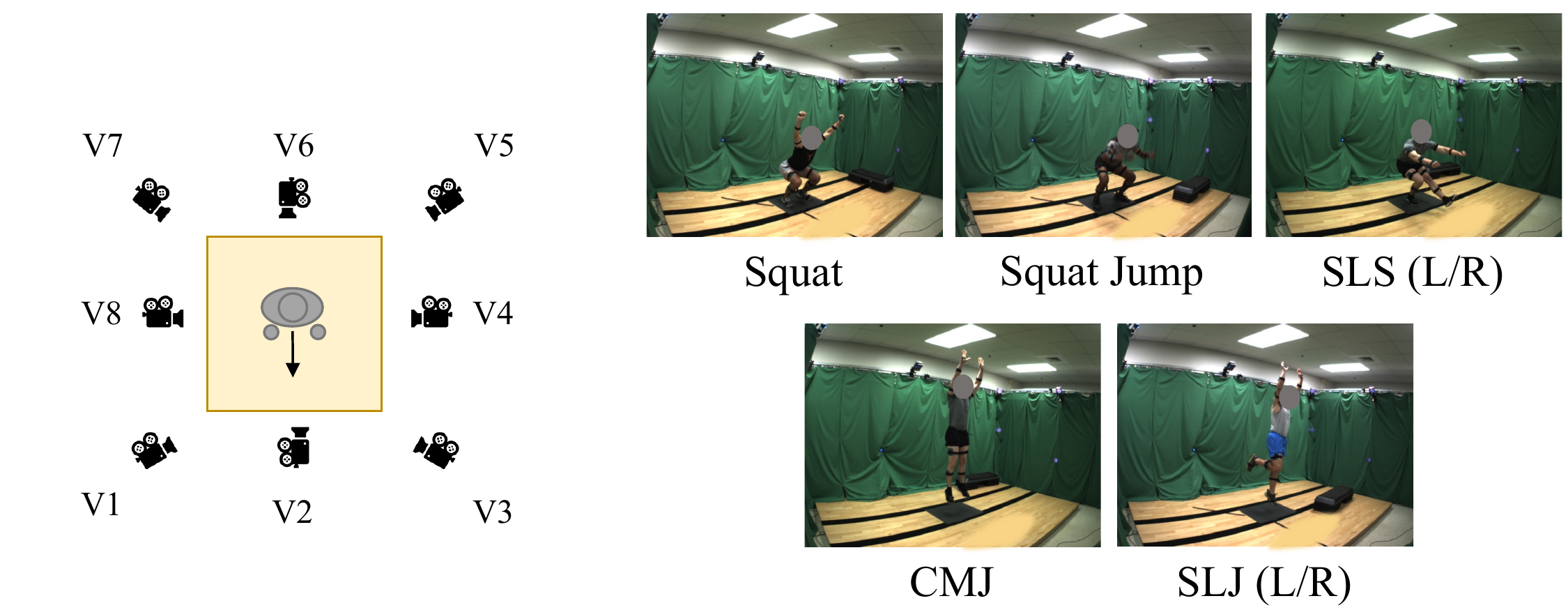}
    \caption{On the left, we show the (eight) camera views used from each trial and on the right, the (six) movements captured in the \swrigrf\ dataset. (Single Leg Squat = SLS, Counter Movement Jump = CMJ, Single Leg Jump = SLJ).}
    \Description{TBD}
    \label{fig:dataset_figure}
\end{figure*}

We introduce the \swrigrf\ dataset, consisting of eight subjects performing five movements for up to three different trials each. These movements are Counter Movement Jump (CMJ), Squat Jump, Squats, Single Leg Squat (SLS) and Single Leg Jump (SLJ) for both left and right legs. We show the camera views and screenshots of these movements in Fig. \ref{fig:dataset_figure}. To promote repeatability and consistency between all experiments, we train with six subjects while we reserving two for validation.
The subjects are fitted with $47$ reflective markers and there are two force plates to capture forces from the left and right foot.
The average mass for training subjects is $88.37 \pm 12.42$ kg and $87.91 \pm 10.74$ kg including validation. 
There are $124$ trials in training and $44$ trials in validation, given eight camera views for each trial there are effectively $1,344$ videos.
This gives us $227,640$ frames in training and $71,400$ frames in validation (including multiple views).
The video data is captured using eight cameras at $50$-Hz, with time-synchronized force plate data recorded at $600$-Hz, and mocap data sourced from a $16$ camera 
marker-based system captured at $100$-Hz. 
We generate 2D pose COCO detections from the video data using a pre-trained Detectron \cite{wu2019detectron2} model, and then compose a pseudo-ground truth 3D pose label from the multiple views.
With intrinsic and extrinsic parameters ($K, T, R$ matrices) of the cameras, we can perform triangulation using a RANSAC-based algorithm to find the best point from all set of observations (viewpoints).
The COCO 2D, 3D detections from all videos and time-synchronized force plate data are publicly available at \url{https://github.com/MichiganCOG/ForcePose}.

\subsection{LAAS Parkour Dataset}
The LAAS Parkour dataset \cite{li2019estimating} is a publicly available dataset of five subjects performing four parkour movements: kong vault, safety vault, pull-up, and muscle-up.
Each subject is fitted with $16$ mocap markers, with force sensors capturing contact forces from the hands (L. Hand, R. Hand) and feet (L. Sole, R. Sole). Like prior work, a mass of $74.6$ kg is assumed for all subjects. For fair comparison with existing work, and with only a single camera view available, we stick to 2D pose detections for these experiments. 
There are $28$ video in total, with an average of $83$ frames per video.
As we are only interested in GRFs, we report our performance on the L. Sole and R. Sole forces.

\section{Method}
We first propose a new loss function (\newloss) for learning GRFs, weighing significant regions heavier in the regressed curve. And we include an additional metric (\newmetric) to measure the error of impact peak forces, comparing the distances between $k$ extrema points. RMSE and \newmetric\ distance metrics together provide a better picture of the characteristics of each prediction describing: (1) how well the overall force curve fits and (2) how closely peak magnitudes are approximated.
Then we describe our process for training the GRF task on a transformer via pre-training and multi-task learning on the 2D-to-3D HPE subtask.

\subsection{Losses and Metrics}
Direct regression of forces using mean-squared error (MSE) as a loss function is the standard approach for learning to approximate ground truth signals in virtually all regression tasks. And for measuring the best fit, root-mean-square error (RMSE) is most often used.
MSE loss and RMSE metric tends to produce or capture a smoothed or averaged approximation of the ground truth signal.
However, in GRF prediction, the instantaneous impact peaks of the force signal are more important than the averaged accuracy across the entire sequence. Therefore we introduce \newloss\ as a substitute for MSE loss, shown below as
\begin{align}
    \mathcal{L}_f = \sum_T w_t \cdot \frac{1}{|\mathbf{F}_{\delta_t}|} \Vert \mathbf{F}_{\delta_t} - \hat{\mathbf{F}}_{\delta_t} \Vert_2^2. \label{eq:loss_force}
\end{align} 
%
Gated-MSE can be viewed as a linear combination of weighted MSE at $T$ thresholds.
We define $\mathbf{F}_{\delta_t} = \{F_1, F_2, \cdots, F_f; \forall |F_f| < \delta_t\}$, which serves as an indexed array for all elements in the ground truth signal $\mathbf{F}$, with an absolute value below the threshold $\delta_n$.
In practice, $T$ serves as a hyper-parameter where each threshold belongs to the sequence $\delta = [0, 1, 5, 10, 15, ...]$ in terms of N/kg. The summed loss is equivalent to the MSE when $T=1$ and we weigh each contribution with $w_t = 1/T$, weighted by the number of thresholds.

RMSE is a useful measure for how well the overall force curve fits, but lacks insight in how closely peak magnitudes are approximated. We use \newmetric\ to provide this measurement.
We extract $k$ extrema ($k$ peaks) and their temporal locations from the forces ($\mathbf{F}$ and $\mathbf{\hat{F}}$), then we compute \newmetric\ using averaged Euclidean distance, $\frac{1}{k} \sum_k \sqrt{x_i^2 + y_i^2}$, along each axis. Here, $x$ represents the temporal distance and $y$ the magnitude difference. 
This additional metric grades the matching capabilities to the high and lowest magnitudes of forces, in addition to a good fit.

\subsection{Predicting Ground Reaction Forces}
In our task, we take an input sequence, $X \in \mathbb{R}^{f \times (J \cdot C)}$, of 2D ($C=2$) or 3D ($C=3$) poses and predict the three-dimensional output force $\mathbf{F}$ at each frame. Here $f$ is the number of frames, $J$ represents the number of joints for the human pose, and $C$ is the number of channels.
We are predicting the magnitude forces on two force plates, so we use a 6D vector representation: $\mathbf{F} = [F_{x1}, F_{y1}, F_{z1}, F_{x2}, F_{y2}, F_{z2}]$. Here $F_x, F_z$ represent the horizontal shear forces, $F_y$ the vertical force, and $1,2$ are for left and right force plates, see Fig. \ref{fig:teaser} for reference.

For training, we use a mini-batch of sliding windows and predict the force at the center frame $\lfloor f/2 \rfloor$. Forces are normalized by subject mass, hence units are in $N/kg$. But we report results in Newtons, multiplying the predicted value with the average training subject mass.
We perform evaluation using \newmetric\ and average RMSE losses across sequences to obtain an Average Sequence Loss shown as:
\begin{align}
    \mathcal{L}_{grf} = \frac{1}{|V|}\sum_{v} \frac{1}{|Cam|} \sum_{cam} \sqrt{\frac{1}{N} \sum_i \Vert \mathbf{F}_i - \hat{\mathbf{F}_i} \Vert^2_2}.
\end{align}

We average the RMSE across all camera views, $Cam$, within a sequence and then average across all video sequences, $V$.

\begin{figure*}[t]
  \centering
  \includegraphics[width=0.95\linewidth]{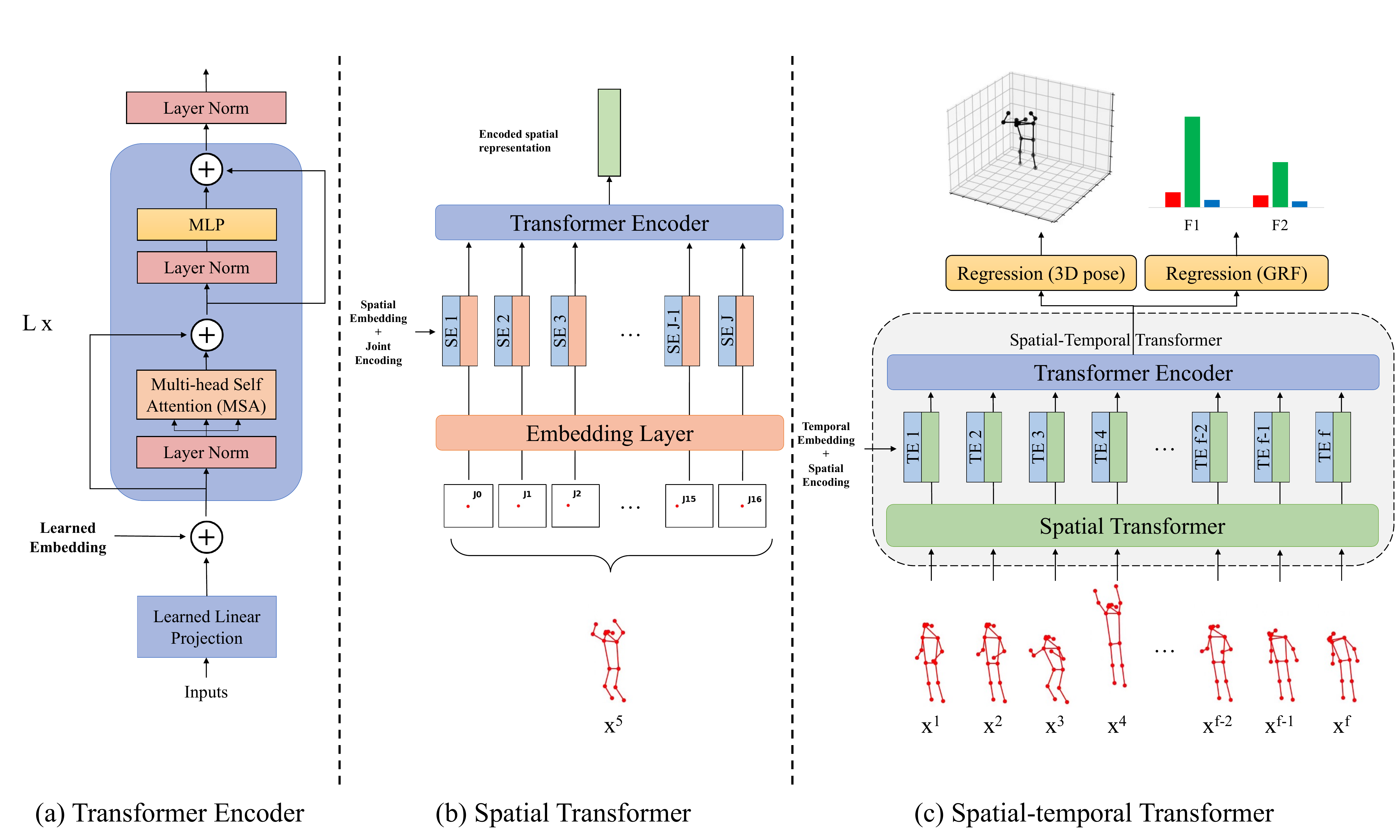}
  \caption{We show a brief overview of (a) the vanilla transformer encoder and the Spatial-Temporal architecture (b)-(c) \cite{zheng20213d} we used in our experiments for this work. This is composed of (b) a spatial encoder and (c) a temporal encoder. The input is a sequence of 2D poses and the output is a prediction of the 3D pose and corresponding GRF at the center of that sequence.}
  \Description{TBD}
  \label{fig:main}
\end{figure*}

\subsubsection{Transformer Encoder}
We use a transformer encoder model to predict the forces, following the spatial-temporal architecture proposed by \cite{zheng20213d}.
Inputs are tokenized at each time step, so a spatial transformer attends to intra-pose coordinates while a temporal transformer attends to inter-pose embeddings as shown in Fig. \ref{fig:main} (b)-(c).
Like standard transformers, first each input is embedded using a learned linear projection layer, $\mathbf{E} \in \mathbb{R}^{(J \cdot C) \times D}$ summed with a learned positional embedding, $\mathbf{E}_{pos} \in \mathbb{R}^{f \times D}$, as shown in Eq. \ref{eq:pos_encoding}. Here $D$ is the dimension of the embedding layer.
Within the encoder, we then pass this embedding through $L$ Multi-head Self Attention (MSA) layers \cite{vaswani2017attention} and $L$ multilayer perceptrons (MLP) with layer normalizations (LN) 
and skip connections \cite{baevski2018adaptive,wang2019learning} in-between, as shown in Eqs. \ref{eq:msa} and \ref{eq:mlp}.
The final output, $Y$, is the $L$th layer output of the encoder after layer normalization.

The operations of the transformer encoder are shown below and in Fig. $\ref{fig:main}$ (a):

\begin{align}
    Z_0 &= [\mathbf{x}^1\mathbf{E};\mathbf{x}^2\mathbf{E}; \cdots ;\mathbf{x}^f\mathbf{E}] + \mathbf{E}_{pos} \label{eq:pos_encoding}\\
    Z_\ell^\prime &= MSA(LN(Z_{\ell-1})) + Z_{\ell-1}, &\ell = 1 \dots L \label{eq:msa}\\
    Z_\ell &= MLP(LN(Z_{\ell-1}^\prime)) + Z_{\ell-1}^\prime, &\ell = 1 \dots L \label{eq:mlp}\\
    Y &= LN(Z_L) \label{eq:ln}
\end{align}

Eqs. \ref{eq:pos_encoding}-\ref{eq:ln} are repeated sequentially for both the spatial and temporal transformer.
The temporal dimension of the final output $Y \in \mathbb{R}^{f \times (J \cdot C)}$ is reduced using a 1D convolution layer as learned weighted averaging, and is passed through a single MLP for the end task prediction.

\subsubsection{Pre-training and Multi-task Learning}
Optimizing the performance of a transformer typically requires a lot of training data. 
To leverage data from a different task, we tightly couple 2D-to-3D HPE and GRF prediction, both benefiting from an implicit 3D representation of the input 2D pose.
Pre-training on the 2D-to-3D HPE task will provide good initial weights and converge to a better solution on the GRF task, especially in situations with scarce amounts of data.
The 2D-to-3D HPE task learns to estimate a 3D pose, $\mathbf{y} \in \mathbb{R}^{J \cdot C}$, at the center of an input sequence.

The loss function is MPJPE (Mean Per Joint Position Error), shown as 
\begin{align}
    \mathcal{L}_{p} = \frac{1}{J}\sum_{k=1}^J \Vert y_k - \hat{y}_k \Vert_2, \label{eq:loss_3d_pose}
\end{align}
between the ground truth and predicted 3D coordinates, averaged over the number of joints. Here $k$ represents each joint index.

After training on 2D-to-3D HPE, we discard the final MLP layer and replace it with one to predict our 6D GRF vector.
We repeat the same steps from the encoder shown in Eqs. \ref{eq:pos_encoding}-\ref{eq:ln}, where the temporal elements of $Y$ are again weighted averaged and passed through a single MLP layer to generate $\mathbf{F}$, $F \in \mathbb{R}^6$.

With multiple views we use multi-task learning (MTL) in place of the ``train then finetune'' paradigm. We optimize the predicted 3D pose concurrently with the 6D ground reaction force (with a separate MLP head), reducing the total training time while retaining solvability of the 2D-to-3D subtask.
Our loss function includes both the 3D pose label and the force vector as a label for supervision. This is directly a summation of Eq. \ref{eq:loss_force} and Eq. \ref{eq:loss_3d_pose} shown as:
\begin{align}
    \mathcal{L}_{mt} &= \mathcal{L}_{f} + \alpha \mathcal{L}_{p}.
\end{align}
Here $L_f$ is the loss for the GRFs, $L_p$ is for 2D-to-3D HPE, and $\alpha$ is a hyper-parameter. We find $\alpha = 1$ to provide the lowest Average Sequence Loss overall.

\begin{table*}
  \centering
  \caption{Average Sequence Losses and \newmetric\ on the \swrigrf\ dataset, measured in Newtons}
  \begin{tabular}{l l l l l l l}
    \toprule
    Pose Format & Input & Prediction Network & RMSE (N) & 1-peak (N) & 3-peak (N) & 5-peak (N) \\
    \midrule
    - & Movement Category & Naive baseline (GRF Exemplar) & 153.84 & 585.28 & 327.35 & 247.86 \\
    COCO & Ankle Keypoints & Naive baseline ($\mathbf{F} = m \ast \mathbf{a}$) & 206.58 & 624.62 & 393.92 & 317.74 \\
    \hline
    \hline
    Mocap markers & Triangulated 3D keypoints & LSTM & \textbf{77.04} $\pm 0.51$ & 322.10 & 199.20 & 165.09 \\
    Mocap markers & Triangulated 3D keypoints & LSTM  (\newlossshort, T=2)& 77.74 $\pm 0.43$ & \textbf{313.34} & 193.60 & 160.19 \\
    Mocap markers & Triangulated 3D keypoints & Transformer & 78.15 $\pm 1.00$ & 373.48 & 217.58 & 175.92 \\
    Mocap markers & Triangulated 3D keypoints & Transformer (\newlossshort, T=2) & 83.50 $\pm 2.97$ & 328.21 & \textbf{193.11} & \textbf{153.72} \\
    \hline 
    COCO  & Triangulated 3D keypoints & LSTM & 81.12 $\pm 1.52$ & 347.46 & 209.14 & 172.72 \\
    COCO  & Triangulated 3D keypoints & LSTM (\newlossshort, T=2) & 82.91 $\pm 1.95$ & \textbf{317.63} & 191.62 & 158.04 \\
    COCO  & Triangulated 3D keypoints & Transformer & \textbf{77.90} $\pm 1.66$ & 340.95 & 199.38 & 157.79 \\
    COCO  & Triangulated 3D keypoints & Transformer (\newlossshort, T=2) & 79.36 $\pm 2.19$ & 336.59 & \textbf{186.43} & \textbf{148.35} \\
    \hline
    COCO  & 2D keypoints & LSTM &  98.14 $\pm 1.81$ & 358.17 & 219.51 & 184.02\\
    COCO  & 2D keypoints & LSTM (\newlossshort, T=2) &  103.02 $\pm 2.88$ & 302.73 & 192.44 & 163.26\\
    COCO  & 2D keypoints & Transformer & \textbf{77.95} $\pm 0.36$ & 321.19 & 190.14 & 154.03 \\
    COCO  & 2D keypoints & Transformer (\newlossshort, T=2) & 83.20 $\pm 3.80$ & \textbf{285.72} & \textbf{171.71} & \textbf{139.23} \\
    \bottomrule
  \end{tabular}
  \label{tab:rmse}
\end{table*}

\section{Experiment Details} \label{sec:experiments}
We experiment with three types of keypoint inputs: 2D, 3D (mocap), and 3D (triangulated). 2D keypoints are generated using a pre-trained Faster-RCNN (ResNet-101) network from Detectron \cite{wu2019detectron2}. We use the default COCO pose format ($17$ keypoints) with no additional finetuning on the detector.
3D keypoints from motion capture (mocap) physical markers are unique to each dataset, hence they do not have a one-to-one correspondence with the COCO format.
With multiple camera views, we construct a triangulated 3D pose from 2D detections as a pseudo-ground truth 3D pose label. These labels are directly associated with their corresponding COCO keypoints.

When training the LSTM baseline, like \cite{morris2021predicting}, we implement a simple bi-directional network with a hidden dimension of $64$-$1024$ (selected through hyper-parameter search), and an MLP encoding layer with a hidden dimension of $256$. We also vary the input sequence length for the network and report only the hyper-parameters leading to the lowest Average Sequence Loss.
The LSTM networks are trained for $40$ epochs using an Adam optimizer, a learning rate of $1e^{-4}$, and a batch size of $64$. 
We use the transformer architecture PoseFormer \cite{zheng20213d} as the base model for our experiments in predicting GRFs from 2D and 3D inputs. For the Transformer encoder, we retain many of the default settings using an embedding dimension $D = 32$, $8$ attention heads, and $4$ depth layers.
From this point, the training details change slightly as we optimize for each input data type and target task, typically until overfitting occurs.
3D (mocap) inputs are trained for $25$ epochs and a learning rate of $4e^{-4}$, 3D (triangulated) inputs for $50$ epochs and a learning rate of $4e^{-4}$,
and 2D inputs for $50$ epochs with a learning rate of $4e^{-4}$. 
When pre-training on the 2D-to-3D HPE, we run our model for up to $100$ epochs and a learning rate of $4e^{-6}$.
All training was conducted with an exponentially decaying learning rate, decaying by a factor of $0.99$ every epoch, and a static batch size of $512$.
We include horizontal flipping of the poses as data augmentation.

\section{Results} \label{sec:results}

\subsection{\swrigrf\ }
We show results on the \swrigrf\ dataset, measuring performance using RMSE for Average Sequence Losses and \newmetric\ as shown in Table \ref{tab:rmse}. 
Each experiment is run for three trials and we report the mean and standard deviation errors.
As mentioned in Sec. \ref{sec:experiments}, we conduct training with three types of keypoints: 3D (mocap), 3D (triangulated), and 2D.
The mocap markers represent the industry standard, hence should result in the lowest error (\textit{i.e.} a soft upper bound). The triangulated 3D keypoints (from 2D detections) require at least two viewpoints and the 2D keypoints require only a single view.

We include two naive baselines for comparison. The first generates an exemplar force profile by averaging the ground truth force of each movement on the training set. But due to high inter-subject variance in force magnitudes and motion timing, the exemplars often do not resemble the typical force profiles.
The second baseline uses Newton's Second Law, $F = m \ast a$, to estimate the total force. With the known frame rate, we estimate the acceleration of the subject using the triangulated 3D ankle keypoint positions. This method is very susceptible to noise and world coordinate scaling errors, and single leg movements introduce difficulties in the computation.
These baselines show very high error on \newmetric\ and an overall worse fit in terms of RMSE.

We compare the performance of the LSTM and transformer architectures across different keypoint input types and then with our \newloss\ during training.
The LSTM initially outperforms the transformer on mocap markers, but as the keypoints get noisier (less perfect) and data samples increases with multiple views, the transformer architecture outperforms an LSTM. This difference is much greater on 2D keypoints, with $20$ N lower error than the LSTM, a reduction of 
$20.5\%$.
The most significant result here is that the performance between the mocap markers and detected COCO keypoints are quite competitive. This highlights that physical markers can be forgone in lieu of pose detections, which are easier to produce and readily available. 
Next, we analyze the impact of our loss across these experiments. Our primary goal is to lower the error in impact peaks, as measured by \newmetric.
We note that while we significantly lower the peak errors in the GRF prediction, this generally comes at a cost in RMSE. But we find that this cost is very low about $1-5$ Newtons, while the average decrease in just a single peak ranges from $10-56$ Newtons.
Through a hyper-parameter search, we find $T=2$ provides the best trade-off between RMSE and minimizing $k$-peak errors. We reserve additional details on loss thresholds and range of input receptive fields for the Supplementary Material. 

\begin{figure*}
  \centering
  \includegraphics[width=\linewidth]{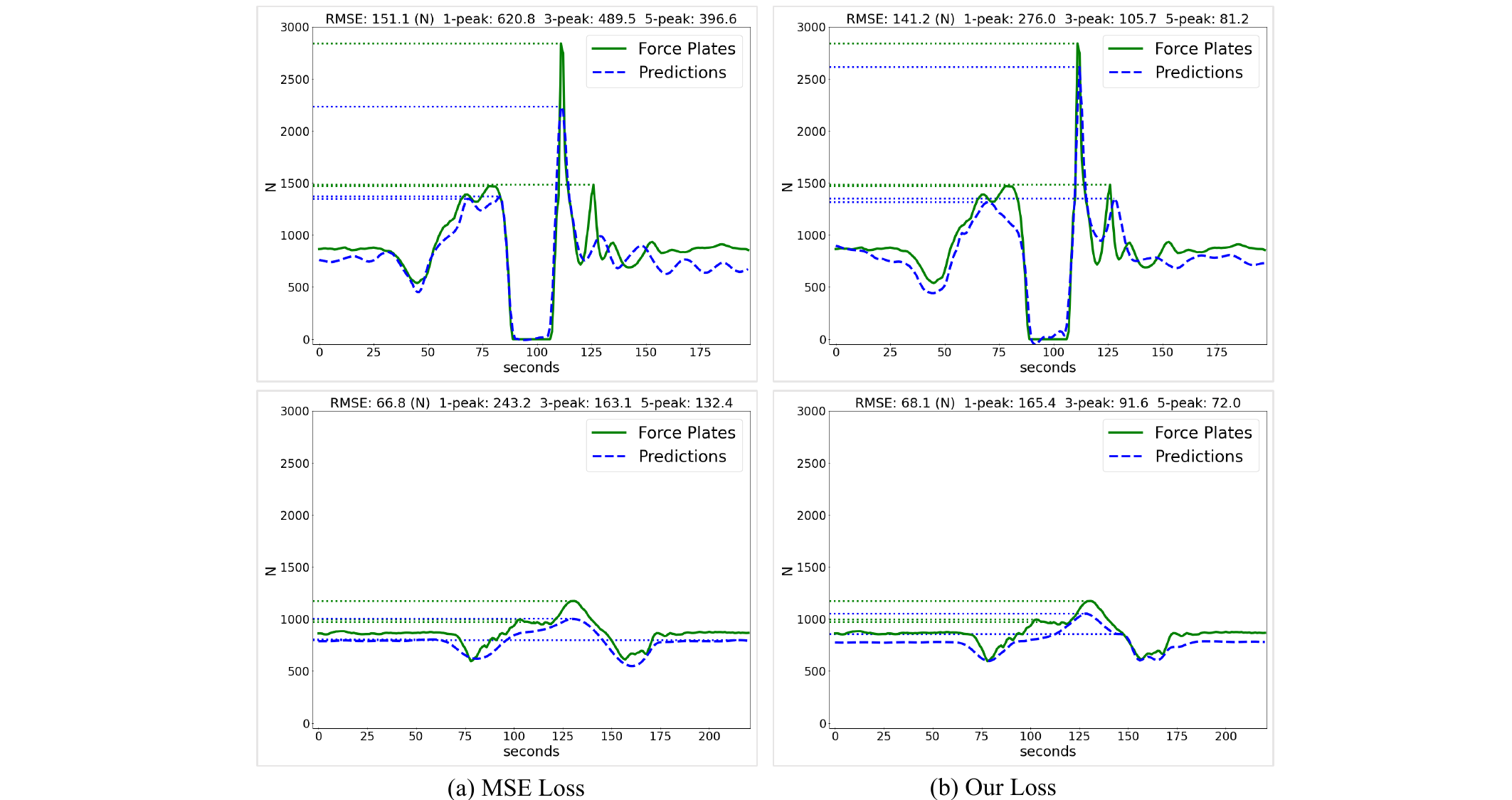}
  \caption{We compare the net GRF outputs on the Single Leg Jump (R) (top row) and Squat (bottom row) trained using the (a) MSE Loss and (b) \newloss. The force plates are shown in green and predicted forces are shown in blue. We show much smaller differences across the \newmetric\ when compared to the ground truth force plates. Horizontal lines mark the detected peaks in each plot, for brevity we only show the top-3. }
  \label{fig:qual_results_02}
\end{figure*}

There is currently no definitive research on acceptable GRF error for downstream biomechanical analysis, as motivated in our introduction. 
However from US Customary Units we can use the conversion $1 \text{ lb.} = 4.448$ N, to better understand the significance of the errors.
For example in Fig. \ref{fig:qual_results_02}, we compare the summed GRF outputs from a model trained on MSE loss versus our new \newloss.
In Fig. \ref{fig:qual_results_02} (a) Single Leg Jump, we decrease the peak magnitude error by $344.8$ N which is approximately $77.5$ lbs. of force and the decrease of the average $5$-peaks is still very high, about $70$ lbs. of force.
This drastic difference shows how vital minimizing these instantaneous errors can be, especially when being used for analysis on the human body.
In the following section, we compare our results on the transformer when pre-training and multi-task learning on a different subtask.

\subsubsection{2D-to-3D subtask} \label{sec:swri_subtask}
We introduce 2D-to-3D HPE as a subtask in training our transformer via pre-training and multi-task learning (MTL), results are shown in Table \ref{tab:transformer_train}. Leveraging the multiple views of \swrigrf, we can use the triangulated 3D poses as pseudo-ground truth labels.
Our MTL implementation performs best across the three training strategies, while training using pre-trained and random weights show similar results.
Optimizing for the 3D pose simultaneously with 3D GRFs provides an advantage because the model maintains a constant three-dimensional representation from the 2D keypoint inputs. This additional constraint puts it on par with the other models that operate directly on 3D keypoints, as shown in Table \ref{tab:rmse}.
On the contrary pre-trained and random weights appear to converge to a similar solution, which we can attribute to the variance and size of the \swrigrf\ dataset. Because the data is large and varied enough for the target GRF task, when trained for enough iterations, the initialization from the pre-training becomes less significant.
We see in Sec. \ref{sec:parkour_results} that when the target dataset is much smaller, the pre-training has a huge impact on the final results.

\begin{table}
    \centering
    \caption{Average Sequence Losses on the \swrigrf\ dataset. We compare results from transformers using 2D-to-3D HPE as a subtask}
    \begin{tabular}{l l l}
        \toprule
         Input & Prediction Network & RMSE (N)\\
         \midrule
         2D Keypoints & Transformer random weights & 80.45 $\pm 1.08$\\ 
         2D Keypoints & Transformer pre-trained (ours) & 80.51 $\pm 0.61$\\ 
         2D Keypoints & Transformer MTL (ours) & \textbf{77.95} $\pm 0.36$\\ 
         %
         \bottomrule
    \end{tabular}
    \label{tab:transformer_train}
\end{table}

\begin{table*}
    \centering
    \caption{Zero-shot learning, RMSE measured in Newtons}
    \begin{tabular}{l l l l l l}
        \toprule
         Method & CMJ & SLS & SLJ & Squat Jump & Squat\\
         \midrule
         Random weights & 117.40 $\pm 2.91$ & 137.76 $\pm 4.82$ & 219.95 $\pm 8.41$ & 104.28 $\pm 4.34$ & 65.77	$\pm 6.22$\\
         Pre-trained weights & 111.62 $\pm 2.45$ & 140.17 $\pm 4.03$ & \textbf{207.01} $\pm 3.68$ & \textbf{104.23} $\pm 1.19$ & 55.15 $\pm 1.29$\\
         Multi-task learning & \textbf{110.29} $\pm 4.59$ & \textbf{119.62} $\pm 1.54$ & 216.92 $\pm 2.57$ & 104.60 $\pm 6.76$ & \textbf{53.37} $\pm 7.79$\\
         \bottomrule
    \end{tabular}
    \label{tab:zero_shot}
\end{table*}

\subsubsection{Zero-shot learning}
With 2D keypoint inputs on the transformer, we measure our ability to generalize to unseen motion using zero-shot learning. We employ leave-one-out cross-validation; training on four movement classes and testing on the unseen class. We compare training with random weights against pre-training and multi-task learning (MTL) on the 2D-to-3D HPE task, reporting our results on Table \ref{tab:zero_shot}. Each column represents the testing (leave out) class.
The advantages of using pre-trained weights or MTL are apparent in all classes except Squat Jump, with Squats having the lowest error overall.
Grouping the classes into squatting and jumping movements, we notice smaller margins between training strategies for jumping, with the exception of single leg.
The jumping motions may appear visually similar, so the additional 2D-to-3D HPE subtask may not provide much more discriminative power, over random weights, on the unseen class.
However, across the variations in the classes, single leg movements are most visually distinct and difficult to generalize to. The subject may position the leg far out front when squatting or further behind them when jumping, deviating far from what has been seen during training.
Interestingly enough, while the performance between pre-training and MTL is generally close, we see a swap between Single Leg Squat and Single Leg Jump.
But when generalizing to newer movements, we find that MTL provides the most advantages and lowest errors across the board. We include additional comparisons and analysis between these training strategies in the Supplementary Material.

\subsection{LAAS Parkour} \label{sec:parkour_results}
\begin{table}
    \centering
    \caption{LAAS Parkour Dataset. Estimation errors of forces (in Newtons). Each subject has an assumed mass of 74.6 kg}
    \begin{tabular}{l l l}
        \toprule
        Method & L. Sole (N) & R. Sole (N)\\
        \midrule 
         Li \etal \cite{li2019estimating} & 144.23 & 138.21 \\
         LSTM & 99.47 & 94.38 \\
         Transformer random weights & 103.48 & 95.08 \\
         Transformer pre-trained (\swrigrf) &  91.47 &  89.67 \\
         Transformer pre-trained (H36m) &  \textbf{83.78} &  \textbf{85.79} \\
         Transformer MTL & 91.74 & 99.78 \\
         \bottomrule
    \end{tabular}
    \label{tab:parkour}
\end{table}

We report results on the LAAS Parkour dataset, shown in Table \ref{tab:parkour}. 
Given the small dataset size, we train and evaluate using leave-one-out cross-validation by subject across all movements. For comparability with other GRF experiments, we only predict the contact forces for the L. Sole and R. Sole. Following \cite{li2019estimating}, the final accuracy is the average L2 distance across all videos. 
Using the same transformer we perform training on random weights, then finetuning from pre-trained weights and multi-task learning on the 2D-to-3D HPE task (\swrigrf\ and Human3.6m \cite{ionescu2013human3} datasets).
The Human3.6m (H36M) dataset is a large mocap dataset with $3.6$ million frames and $17$ actions but no force data.
We use this pre-training only on LAAS to showcase the impact of much larger datasets when finetuning on smaller datasets. But we maintain COCO detections throughout other experiments for consistency.
There is not a one-to-one mapping between the COCO keypoints and the H36M mocap markers, hence the transformer must learn some keypoint re-mapping during the finetune stage.  

We compare to prior work by Li \etal \cite{li2019estimating}, they re-project detected 2D joint locations to 3D positions and solve a large-scale trajectory optimization problem to learn contact forces and 3D motions.
We show that the baseline LSTM network outperforms the previous SOTA baseline with a 
$30\%$ decrease in error. 
Motivating the strategy of pre-training on the 2D-to-3D HPE task, we note a $9 - 19\%$ decrease in error between that and a transformer trained from random weights.
 With the LAAS Parkour dataset being very small, unlike the results shown in Sec. \ref{sec:swri_subtask}, pre-training on 2D-to-3D HPE demonstrates a tremendous advantage from using the \swrigrf\ and Human3.6m datasets.
This trend is sustained even with the keypoint mismatch between H36M and COCO pose formats, showing the lowest loss overall.
The transformer training from random weights performs worse than the LSTM baseline, due to the small number of data samples paired with a more complex network architecture.
And MTL training performs well on the L. Sole, but much worse on the R. Sole. We believe this may originate from restrictions in the dataset, which only supports the same single view for all videos.
In $68\%$ of the trials, the subject is facing to the left which may bias the $3$D pose estimation branch of the MTL.

\section{Conclusion} \label{sec:conclusion}
In this work we present a new approach for solving the GRF prediction from video problem. First, we introduced the \swrigrf\ dataset, a large collection of multi-view tracked human motions with paired force plate data.
Then we addressed the minimization of peak impact errors by introducing \newloss\, which drastically reduces peak impact errors at a low cost in RMSE.
We also show, using a transformer architecture, that pre-training and multi-task learning on 2D-to-3D human pose estimation induces better performance when fine-tuning on small datasets and has better generalization to unseen motions.
This work takes a great step towards analyzing the quality of human motions and actions. By estimating these values directly from video we can use them as apart of an automated process for comprehensive analysis.

\begin{acks}
We would like to thank Madan R. Ganesh, Stephen J. Lemmer, and Luowei Zhou for their insightful discussions and input.
This work was supported in part by a UM fellowship and NIH Award 5 R01 HL146619-03.
\end{acks}

\bibliographystyle{ACM-Reference-Format}
\bibliography{acmart}
\end{document}


\title{Supplementary Material: Learning to Estimate External Forces of Human Motion in Video}

\author{Nathan Louis}
\authornote{Completed as part of internship at Southwest Research Institute}
\email{natlouis@umich.edu}
\orcid{0000-0003-4502-6012}
\affiliation{%
  \institution{University of Michigan}
  \city{Ann Arbor}
  \state{Michigan}
  \country{USA}
  \postcode{48109}
}

\author{Jason J. Corso}
\affiliation{%
  \institution{University of Michigan}
  \city{Ann Arbor}
  \state{Michigan}
  \country{USA}
  \postcode{48109}}
\email{jjcorso@umich.edu}

\author{Tylan N. Templin}
\affiliation{%
  \institution{Southwest Research Institute}
  \city{San Antonio}
  \state{Texas}
  \country{USA}}
\email{tylan.templin@swri.org}

\author{Travis D. Eliason}
\affiliation{%
  \institution{Southwest Research Institute}
  \city{San Antonio}
  \state{Texas}
  \country{USA}}
\email{travis.eliason@swri.org}

\author{Daniel P. Nicolella}
\affiliation{%
  \institution{Southwest Research Institute}
  \city{San Antonio}
  \state{Texas}
  \country{USA}}
\email{dnicolella@swri.org}

\renewcommand{\shortauthors}{Louis, et al.}

\maketitle

\section{Hyper-parameters}

\subsection{Input receptive field}
When training all transformer models, we experiment with four input receptive fields on the \swrigrf\ dataset; $9, 27, 43, 81$ frames. As shown in Table \ref{tab:transformer_seq_len}, the longer sequences result in lower average sequence losses across all trained models. We report only the best (lowest) loss in main manuscript, hence all transformer models shown are trained on $81$ frames. 
Across the different training strategies, our multi-task learning (MTL) consistently produces the lowest average sequence loss.

We also include the results over the various receptive fields for zero-shot learning in Table \ref{tab:zero_shot_full}, our results are consistent with the transformer models trained on all classes.

\begin{table*}
\centering
\caption{Sweep input receptive field}
    \begin{tabular}{l l l l l}
    \toprule
    Training Strategy & 9 frames         & 27 frames        & 43 frames        & 81 frames \\
    \midrule
    Transformer random weights       & 92.67 $\pm$ 1.97 & 85.82 $\pm$ 1.45 & 80.91 $\pm$ 0.51 & 80.45 $\pm$ 1.08 \\
    Transformer pre-trained & 92.71 $\pm$ 1.01 & 84.00 $\pm$ 0.65 & 80.99 $\pm$ 0.39 & 80.51 $\pm$ 0.61 \\           
    Transformer MTL & \textbf{90.69} $\pm 0.80$ & \textbf{81.73} $\pm 0.59$ & \textbf{78.51} $\pm 0.55$ & \textbf{77.95} $\pm 0.36$ \\
    \bottomrule
    \end{tabular}
\label{tab:transformer_seq_len}
\end{table*}

\begin{table*}
    \centering
    \caption{Zero-shot learning, sweep input receptive field (single run)}
    \begin{tabular}{l l  l  l  l  l  l}
        \toprule
         9 frames & CMJ & SLS & SLJ & SJ & Squat & Average\\
         \midrule
         Random Weights & \textbf{139.69} & 105.90 & 228.47 & 129.34 & \textbf{78.88} & 136.45 \\
         Pre-trained Weights & 146.86 & 111.72 & \textbf{217.26} & 129.96 & 101.75 & 141.51 \\
         Multi-task Learning & 140.42 & \textbf{102.07} & 223.66 & \textbf{127.52} & 87.69 & \textbf{136.27} \\
        \\
        27 frames \\
        \midrule
        Random Weights & 132.00 & \textbf{97.19} & 222.82 & 120.63 & 72.59 & 129.04 \\
        Pre-trained Weights & 126.92 & 111.40 & 224.67 & \textbf{115.28} & 73.47 & 130.34 \\
        Multi-task Learning & \textbf{120.80} & 117.44 & \textbf{217.74} & 116.91 & \textbf{68.85} & \textbf{128.35} \\
        \\
        43 frames \\
        \midrule
        Random Weights & 114.37 & \textbf{104.99} & 215.66 & \textbf{104.34} & 76.14 & 123.10 \\
        Pre-trained Weights & 109.96 & 116.25 & 215.90 & 105.74 & 68.22 & 123.21 \\
        Multi-task Learning & \textbf{109.67} & 109.05 & \textbf{211.29} & 108.70 & \textbf{67.53} & \textbf{121.25} \\
        \bottomrule
    \end{tabular}
    \label{tab:zero_shot_full}
\end{table*}

\subsection{Loss function thresholds}
The \newloss\ loss includes a threshold hyper-parameter, $T$, in the loss function:
\begin{align}
    \mathcal{L}_f = \sum_T w_t \cdot \frac{1}{|\mathbf{F}_{\delta_t}|} \Vert \mathbf{F}_{\delta_t} - \hat{\mathbf{F}}_{\delta_t} \Vert_2^2. \label{eq:loss_force}
\end{align}
$T$ specifies the number of terms to sum in the total loss, from the sequence $\delta = [0, 1, 5, 10, 15, ...]$ (N/kg). Again, $T=1$ is equivalent to the MSE and $w_t = 1/T$. In Table \ref{tab:sweep_threshold} we show the results of sweeping through the $T=1...3$.
Across the trained models, we find $T=2$ provides the greatest reduction in \newmetric\ while minimizing the additional cost in RMSE fit.

\begin{table*}
    \centering
    \caption{Sweep threshold, T, in \newloss\ loss}
    \begin{tabular}{l l l l l l}
    \toprule
         Prediction Network & RMSE (N) & 1-peak (N) & 3-peak (N) & 5-peak (N) \\
         \midrule
         Transformer random weights & \textbf{80.45} $\pm$ 1.08 & 333.12 & 196.67 & 155.83 \\
         Transformer random weights (\newlossshort, T=2) & 83.95 $\pm$ 3.80 & \textbf{282.92} & \textbf{174.95} & \textbf{143.83} \\
         Transformer random weights (\newlossshort, T=3) & 89.68 $\pm$ 4.80 & 296.72 & 177.99 & 147.04 \\
         \midrule
         Transformer MTL & \textbf{77.95} $\pm$ 0.36 & 321.19 & 190.14 & 154.03 \\
         Transformer MTL (\newlossshort, T=2) & 84.95 $\pm$ 4.35 & 285.10 & 170.52 & 140.83 \\
         Transformer MTL (\newlossshort, T=3) & 86.23 $\pm$ 1.11 & \textbf{277.25} & \textbf{170.10} & \textbf{139.57} \\
         \midrule
         Transformer pre-trained & \textbf{80.51} $\pm$ 0.61 & 320.41 & 189.99 & 151.40 \\
         Transformer pre-trained (\newlossshort, T=2) & 83.20 $\pm$ 0.39 & \textbf{285.72} & 171.71 & 139.23 \\
         Transformer pre-trained (\newlossshort, T=3) & 84.58 $\pm$ 1.87 & 293.20 & \textbf{170.15} & \textbf{134.57} \\
         \bottomrule
    \end{tabular}
    \label{tab:sweep_threshold}
\end{table*}
